\documentclass[11pt]{article}
\usepackage{amssymb}
\usepackage{multirow}
\usepackage{amsmath}
\usepackage{amsfonts}
 \usepackage{pifont}
\usepackage{amsmath,amsfonts,amssymb,theorem,graphicx,listings,txfonts}
\usepackage{graphics}
 \usepackage{caption2}
\usepackage{flafter}
 \usepackage{indentfirst}
\usepackage{epsfig}
  \usepackage{mathrsfs}
  \usepackage{float}
\usepackage{subfigure}
 \usepackage{cite}
\newtheorem{theorem}{Theorem}[section]

\newtheorem{proposition}[theorem]{Proposition}

\newtheorem{definition}[theorem]{Definition}
\newtheorem{algorithm}[theorem]{Algorithm}

\theoremstyle{example}
\newtheorem{example}[theorem]{Example}
\theoremstyle{programme}

\theoremstyle{property}

\theoremstyle{problem}

\title{Related family-based attribute reduction of covering information systems when varying attribute sets}

\author
{Guangming Lang$^{1,2,3}$ 
\thanks{Corresponding author:\quad langguangming1984@tongji.edu.cn.
\newline\mbox{}\hspace{0.55cm}
E-mail address: langguangming1984@tongji.edu.cn. }\hspace{1cm}\\
\small {$^{1}$ School of Mathematics and Statistics, Changsha University of Science and Technology}\\
\small {Changsha, Hunan 410114, P.R. China}\\
\small {$^{2}$ Department of Computer Science and Technology, Tongji University}\\
\small {Shanghai 201804, P.R. China}\\
\small {$^{3}$ The Key Laboratory of Embedded System and Service Computing, Ministry of Education, Tongji University}\\
\small {Shanghai 201804, P.R. China}\\
\small {$^{4}$ College of Mathematics and Econometrics, Hunan University}\\
\small {Changsha, Hunan 410004, P.R. China}}
\date{}

\begin{document}
\maketitle \baselineskip=17pt
\begin{center}
\begin{quote}
{{\bf Abstract.}
In practical situations, there are many dynamic covering information systems with variations of attributes, but there are few studies on related family-based attribute reduction of dynamic covering information systems.
In this paper, we first investigate updated mechanisms of constructing attribute reducts for consistent and inconsistent covering information systems when varying attribute sets by using related families. Then we employ examples to illustrate how to compute attribute reducts of dynamic covering information systems with variations of attribute sets.
Finally, the experimental results illustrates that the related family-based methods are effective to perform attribute reduction of dynamic covering information systems when attribute sets are varying with time.

{\bf Keywords:} Attribute reduction; Covering information system; Dynamic information system; Related family; Rough sets
\\}
\end{quote}
\end{center}
\renewcommand{\thesection}{\arabic{section}}

\section{Introduction}

Covering rough set theory, pioneered by Zakowski\cite{Zakowski} in 1983, has become an useful mathematical tool for dealing with uncertain and imprecise information in practical situations. As a substantial constituent of granular computing, covering-based rough set theory has been applied to many fields such as feature selection and data mining without any prior knowledge. Especially, covering rough set theory is being attracting more and more attention in the era of artificial intelligence, which will provide powerful supports for the development of data processing technique.

Many researchers\cite{Bonikowski,Leung1,Li5,Liu3,Pomykala,Chen3,Hu2,Hu3,Huang1, Huang2,Huang3,Lang2,Li1,Li2,Li3,Li4,Liu1,Liu4,Ma1,Tsang,Tan1,Wang5,Wu1,Xu2,Yang1,Yang2,
Yang3,Yang4,Yang6,Yao1,Yao2,Yao3,Zhang2,Zhang3,Zhang4,Zhu1,Zhu2,Zhu3} have studied covering-based approximations of sets. For example, After Zakowski extended Pawlak's rough set theory\cite{Pawlak1}, the second and third types of covering rough set models are proposed by Pomykala\cite{Pomykala} and Tsang et al.\cite{Tsang}, respectively. Hu et al.\cite{Hu2} proposed matrix-based approaches for dynamic updating approximations in multigranulation rough sets.
Lang et al.\cite{Lang2} presented incremental approaches to computing the second and sixth lower and upper approximations of sets in dynamic covering approximation spaces. Luo et al.\cite{Luo1} presented the updating properties for dynamic maintenance of approximations when the criteria values in set-valued decision systems evolve with time. Tan et al.\cite{Tan1} introduced matrix operations for computing the positive regions of covering decision information systems. Wang et al.\cite{Wang5} transformed the set approximation computation into products of the type-1 and type-2 characteristic matrices and the characteristic function of the set in covering approximation spaces. Yang et al.\cite{Yang2}
investigated a fuzzy covering-based rough set model and its generalization over fuzzy lattice. Yang et al.\cite{Yang3} discussed the relationship among these approximation operators and investigated knowledge reduction about approximation spaces of covering generalized rough sets. Yang et al.\cite{Yang4} provided related family-based methods for attribute reduction of covering information systems.
Yao et al.\cite{Yao2} classified all approximation operators into element-based approximation operators, granule-based approximation operators, and subsystem-based approximation operators.
Zhang et al.\cite{Zhang2} updated the relation matrix to compute lower and upper approximations with dynamic attribute variation in set-valued information systems.
Zhu\cite{Zhu1} provided an approach
without using neighborhoods for studying covering rough sets based on neighborhoods. Zhu\cite{Zhu2} investigated relationship among basic concepts in covering-based rough sets.

Knowledge reduction of dynamic information systems\cite{Cai, Chen1,Chen2,Hu1,Lang1,Lang3,Li6,Li2,Liang1,Liu2,Luo1,Luo2,Qian1,Qian2,Sang,Shu1,Shu2,
Tan2,Wang1,Wang2,Wang3,Wang4,Xu1,Yang5,Zhang1,Zhang4} has attracted more attention. For example, Chen et al.\cite{Chen1} employed an incremental manner to update minimal elements in the discernibility matrices at the arrival of an incremental sample. Lang et al.\cite{Lang1} focused on knowledge reduction of dynamic covering information systems with variations of objects using the type-1 and type-2 characteristic matrices.
Li et al.\cite{Li2} discussed the principles of updating $P$-dominating sets and $P$-dominated sets when some attributes are added into or deleted from the attribute set $P$.
Liang et al.\cite{Liang1} proposed a group incremental
approach to feature selection applying rough set technique.
Luo et. al\cite{Luo2} provided efficient approaches for updating probabilistic approximations with incremental objects.
Qian et al.\cite{Qian1} focused on attribute reduction for sequential three-way decisions under dynamic granulation.
Wang et al.\cite{Wang1} investigated efficient updating rough approximations with multi-dimensional variation of ordered data.
Xu et al.\cite{Xu1} proposed a three-way decisions model with probabilistic rough sets for stream computing. Yang et al.\cite{Yang5}
investigated fuzzy rough set based incremental attribute reduction from dynamic data with sample arriving. Zhang et al.\cite{Zhang1} provided incremental approaches for computing the lower and upper approximations with dynamic attribute variation in set-valued information systems.
Zhang et al.\cite{Zhang4} provided a parallel matrix-based method for computing approximations in incomplete information systems.

In practical situations, there are many types of covering information systems such as incomplete information systems and set-valued information systems. Especially, covering-based information systems are varying with the time, and knowledge reduction of dynamic covering information systems is a significant challenge of covering-based rough sets. So far, there are many methods for attribute reduction of covering information systems. Especially, related family-based methods proposed by Yang\cite{Yang2} are very effective for knowledge reduction of covering information systems, and bridge the gap where the discernibility matrix is not applicable. In practical situations, there are many dynamic covering information systems with variations of object sets, attribute sets, and attribute values. But
there are few researches on knowledge reduction of dynamic covering information systems using related families.
The purpose of this paper is to investigate knowledge reduction of dynamic covering decision information systems with related families. First,
we study knowledge reduction of consistent covering information systems with variations of attribute sets. Concretely, we construct the related family of dynamic covering information systems based on that of original consistent information systems. We also investigate the relationship between attribute reducts of dynamic covering information systems and that of original consistent information systems. We employ several examples to illustrate how to compute attribute reducts of dynamic covering information systems with related families. Second,
we study knowledge reduction of inconsistent covering information systems with variations of attribute sets. Concretely, we construct the related family of dynamic covering information systems based on that of original inconsistent information systems. We also investigate the relationship between attribute reducts of dynamic covering information systems and that of original inconsistent covering information systems. We employ several examples to illustrate how to compute attribute reducts of dynamic covering information systems with related families. Third, we perform the experiments on data sets downloaded from UCL, and the experimental results illustrates that the related family-based methods are effective for knowledge reduction of dynamic covering information systems with variations of attribute sets.

The rest of this paper is organized as follows: In Section 2, we briefly
review the basic concepts of covering-based rough set theory. In
Section 3, we study updated mechanisms for constructing attribute reducts of consistent covering information systems with variations of attribute sets using related families. In Section 4, we constructed attribute reducts of inconsistent covering information systems when varying attribute sets using related families. 
Concluding remarks and further research are given in Section 5.

\section{Preliminaries}

In this section, we  briefly review some concepts of covering-based rough
sets.

\begin{definition}\cite{Zakowski}
Let $U$ be a finite universe of discourse, and $\mathscr{C}$ a
family of subsets of $U$. Then $\mathscr{C}$ is called a covering of
$U$ if none of elements of $\mathscr{C}$ is empty and
$\bigcup\{C\mid C\in \mathscr{C}\}=U$. Furthermore, $(U,\mathscr{C})$ is referred to as a covering approximation space.
\end{definition}

If $U$ is a finite universe of discourse, and $\Delta=\{\mathscr{C}_{1},\mathscr{C}_{2},...,\mathscr{C}_{m}\}$, where $\mathscr{C}_{i}$ $(1\leq i\leq m)$ is a
covering of $U$, then $(U,\Delta)$ is called a covering information system; if all coverings are classified into conditional attribute-based coverings and decision attribute-based coverings, then $(U,\Delta, D)$ is called a covering decision information system, where $\Delta$ and $D$ denote the conditional attributes and  decision attributes, respectively.

\begin{definition}\cite{Zhu3}
Let $(U,\mathscr{C})$ be a covering approximation
space, and $Md_{\mathscr{C}}(x)=\{K\in \mathscr{C}\mid x\in K\wedge (\forall S\in \mathscr{C}\wedge x\in S \wedge S\subseteq K\Rightarrow K=S)\}$ for $x\in U$. Then $Md_{\mathscr{C}}(x)$ is called the minimal description of $x$.
\end{definition}

The minimal description of $x$ is a set of the minimal elements containing $x$ in $\mathscr{C}$. For a covering $\mathscr{C}$ of $U$, $K$ is a union reducible element of $\mathscr{C}$, $\mathscr{C}-\{K\}$ and $\mathscr{C}$ have the same $Md(x)$ for $x\in U$. If $K$ is a union reducible element of $\mathscr{C}$ if and only if $K\notin Md(x)$ for any $x\in U$, and denote $\mathscr{M}_{\cup\Delta}=\{Md_{\cup\Delta}(x)\mid  x\in U\}$ with respect to a family of coverings $\Delta$.

\begin{definition}\cite{Zhu3}
Let $(U,\mathscr{C})$ be a covering approximation
space, and $Md_{\mathscr{C}}(x)$ the minimal description of $x\in U$. Then the third lower and upper approximations
of $X\subseteq U$ with respect to $\mathscr{C}$ are defined as follows:
\begin{eqnarray*}
CL_{\mathscr{C}}(X)=\cup\{K\in \mathscr{C}\mid K\subseteq X\} \text{ and }
CH_{\mathscr{C}}(X)=\cup\{K\in Md_{\mathscr{C}}(x)\mid x\in X\}.
\end{eqnarray*}
\end{definition}

The third lower and upper approximation operators are typical representatives of non-dual approximation operators for covering approximation spaces. Furthermore, we have $
CL_{\mathscr{C}}(X)=\bigcup\{K\in \mathscr{C}\mid \exists x, $ s.t. $ (K\in Md_{\mathscr{C}}(x))\wedge (K\subseteq X)\}
$ with the minimal descriptions. Especially, we
have $CL_{\cup\Delta}(X)=\cup\{K\in Md_{\cup\Delta}(x)\mid K\subseteq X\} \text{ and } CH_{\cup\Delta}(X)=\cup\{K\in Md_{\cup\Delta}(x)\mid x\in X\}.$
For simplicity, we denote $POS_{\cup\Delta}(X)=CL_{\cup\Delta}(X),BND_{\cup\Delta}(X)$ $=CH_{\cup\Delta}(X)\backslash CL_{\cup\Delta}(X),$ and $NEG_{\cup\Delta}(X)=U\backslash CH_{\cup\Delta}(X).$

\begin{definition}
Let $(U,\Delta, D)$ be a covering information system, where $U=\{x_{1},x_{2},...,x_{n}\}$, $\Delta=\{\mathscr{C}_{1},\mathscr{C}_{2},...,$ $\mathscr{C}_{m}\}$, and $U/D=\{D_{1},D_{2},...,D_{k}\}$. Then

(1) if for $\forall x\in U, \exists D_{j}\in U/D$ and $\exists K\in Md_{\cup\Delta}(y)$ such that $x\in K\subseteq D_{j}$, where $y\in U$, then the decision system $(U,\Delta, D)$ is called a consistent covering decision information system.

(2) if there exists $x\in U$ but $\overline{\exists} K\in \cup\Delta \text{ and } D_{j}\in U/D$ such that $x\in K\subseteq D_{j}$, then the decision system $(U,\Delta, D)$ is called an inconsistent covering decision information system.
\end{definition}

For simplicity, if $(U,\Delta, D)$ is a consistent covering decision information system, then we denote it as $\mathscr{M}_{\cup\Delta}\preceq U/D$; if $(U,\Delta, D)$ is
an inconsistent covering decision information system, then we denote it as $\mathscr{M}_{\cup\Delta}\npreceq U/D$.

\begin{definition}
Let $(U,\Delta, D)$ be a covering information system, where $U=\{x_{1},x_{2},...,x_{n}\}$, $\Delta=\{\mathscr{C}_{1},\mathscr{C}_{2},...,$ $\mathscr{C}_{m}\}$, and $U/D=\{D_{1},D_{2},...,D_{k}\}$. Then

(1) if $POS_{\cup\Delta}(D)=POS_{\cup\Delta-\{\mathscr{C}_{i}\}}(D)$ for $\mathscr{C}_{i}\in \Delta$, where $POS_{\cup\Delta}(D)=\bigcup \{POS_{\cup\Delta}(D_{i})$ $\mid D_{i}\in U/D\}$, then $\mathscr{C}_{i}$ is called superfluous relative to $D$; Otherwise, $\mathscr{C}_{i}$ is called indispensable relative to $D$;

(2) if every element of $P\subseteq \Delta$ satisfying $\mathscr{M}_{\cup P}\preceq U/D$ is indispensable relative to $D$, then $P$ is called a reduct of $\Delta$ relative to $D$.
\end{definition}

By Definition 2.5, we have the following results: if $(U,\Delta, D)$ is a consistent information system, then we have $POS_{\cup\Delta}(D)=U$; if $(U,\Delta, D)$ is an inconsistent information system, then we have $POS_{\cup\Delta}(D)\neq U$.

\begin{definition}
Let $(U,\Delta, D)$ be a covering information system, where $U=\{x_{1},x_{2},...,x_{n}\}$, $\Delta=\{\mathscr{C}_{1},\mathscr{C}_{2},...,$ $\mathscr{C}_{m}\}$,
$\mathscr{A}=\{C_{k}\in \cup \Delta\mid \exists D_{j}\in U/D, $ s.t. $ C_{k}\subseteq D_{j}\}$, and $r(x_{i})=\{\mathscr{C}\in \Delta\mid \exists C_{k}\in \mathscr{A}, $ s.t. $  x_{i}\in C_{k}\in \mathscr{C}\}.$ Then $R(U,\Delta,D)=\{r(x_{i})\mid r(x_{i})\neq \emptyset,x_{i}\in U,\}$ is called the related family of $(U,\Delta, D)$.
\end{definition}

By Definition 2.6, we have the following results: if $(U,\Delta, D)$ is a consistent information system, then we have $r(x)\neq \emptyset$ for any $x\in U$; if $(U,\Delta, D)$ is an inconsistent information system, then we have $r(x)=\emptyset$ for some $x\in U$.

\begin{definition}
Let $(U,\Delta, D)$ be a covering information system, where $U=\{x_{1},x_{2},...,x_{n}\}$, $\Delta=\{\mathscr{C}_{1},\mathscr{C}_{2},...,$ $\mathscr{C}_{m}\}$, and
$R(U,\Delta,D)$ the related family of $(U,\Delta, D)$. Then

(1) $f(U,\Delta,D)=\bigwedge\{\bigvee r(x_{i})\mid r(x_{i})\in R(U,\Delta,D)\}$ is the related function, where $\bigvee r(x_{i})$ is the disjunction of all elements in $r(x_{i})$;

(2) $g(U,\Delta,D)=\bigvee^{l}_{i=1}\{\bigwedge \Delta_{i}\mid \Delta_{i}\subseteq\Delta\}$ is the reduced disjunctive form of $f(U,\Delta,D)$ with the multiplication and absorption laws.
\end{definition}

By Definition 2.7, we have attribute reducts $\mathscr{R}(\Delta,U,D)=\{\Delta_{1},\Delta_{1},...,\Delta_{l}\}$ for $(U,\Delta, D)$. We also present a non-incremental algorithm of computing $\mathscr{R}(U,\Delta,D)$ for covering decision information system $(U,\Delta,D)$ as follows.

\begin{algorithm}(Non-Incremental Algorithm of Computing $\mathscr{R}(U,\Delta,D)$ for Covering Information System $(U,\Delta,D)$)(NIACIS).

Step 1: Input $(U,\Delta, D)$;

Step 2: Construct $POS_{\cup\Delta}(D)=\bigcup \{POS_{\cup\Delta}(D_{i})$ $\mid D_{i}\in U/D\}$;

Step 3: Compute $R(U,\Delta,D)=\{r(x_{i})\mid x_{i}\in U,r(x)\neq \emptyset\}$, where \begin{eqnarray*}
r(x_{i})&=&\{\mathscr{C}\in \Delta\mid \exists C\in \mathscr{A}, \text{ s.t. }  x_{i}\in C\in \mathscr{C}\};\\
\mathscr{A}&=&\{C\in \cup \Delta\mid \exists D_{j}\in U/D, \text{ s.t. } C\subseteq D_{j}\};
\end{eqnarray*}

Step 4: Construct $f(U,\Delta,D)=\bigwedge\{\bigvee r(x_{i})\mid r(x_{i})\in R(U,\Delta,D)\}$;

Step 5: Compute $g(U,\Delta,D)=\bigvee^{l}_{i=1}\{\bigwedge \Delta_{i}\mid \Delta_{i}\subseteq\Delta\}$;

Step 6: Output $\mathscr{R}(\Delta,U,D)$.
\end{algorithm}

The time complexity of Step 2 is $[|U|\ast(\sum_{\mathscr{C}\in\Delta}|\mathscr{C}|),|U|\ast(\sum_{\mathscr{C}\in\Delta}|\mathscr{C}|)\ast |U/D|]$; the time complexity of Step 3 is $[|U|^{2},|U|^{2}\ast(\sum_{\mathscr{C}\in\Delta}|\mathscr{C}|)\ast |U/D|]$; the time complexity of Steps 4 and 5 is $[|U|,|U|\ast(|\Delta|+1)]$. Therefore, the
time complexity of the non-incremental algorithm is very high.

\section{Related family-based attribute reduction of consistent covering information systems with variations of attributes}

In this section, we study related family-based attribute reduction of consistent covering information systems with variations of attributes.

\begin{definition}
Let $(U,\Delta, D)$ and $(U,\Delta^{+}, D)$ be covering information systems, where $U=\{x_{1},x_{2},...,x_{n}\}$, $\Delta=\{\mathscr{C}_{1},
\mathscr{C}_{2},...,\mathscr{C}_{m}\}$, and $\Delta^{+}=\{\mathscr{C}_{1},
\mathscr{C}_{2},...,\mathscr{C}_{m},\mathscr{C}_{m+1}\}$. Then $(U,\Delta^{+}, D)$ is called a dynamic information system of $(U,\Delta, D)$.
\end{definition}

\noindent\textbf{Remark:} We take $(U,\Delta, D)$ as a consistent covering information system in Definition 3.1. We also notice that the dynamic covering information system $(U,\Delta^{+}, D)$ is consistent when adding the covering $\mathscr{C}_{m+1}$ into $(U,\Delta,D)$. In practical situations, there are many dynamic covering information systems, and we only discuss consistent covering information systems with variations of attributes in this section.

\begin{example}
Let $(U,\Delta, D)$ and $(U,\Delta^{+}, D)$ be covering information systems, where $U=\{x_{1},x_{2},...,x_{8}\}$, $\Delta=\{\mathscr{C}_{1},
\mathscr{C}_{2},\mathscr{C}_{3},\mathscr{C}_{4},\mathscr{C}_{5}\}$, $\Delta^{+}=\{\mathscr{C}_{1},
\mathscr{C}_{2},\mathscr{C}_{3},\mathscr{C}_{4},\mathscr{C}_{5},\mathscr{C}_{6}\}$, and $U/D=\{\{x_{1},x_{2},x_{3}\},\{x_{4},x_{5},x_{6}\},\{x_{7},x_{8}\}\}$, where
\begin{eqnarray*}
\mathscr{C}_{1}&=&\{\{x_{1},x_{2}\},\{x_{2},x_{3},x_{4}\},\{x_{3}\},
\{x_{4}\},\{x_{5},x_{6}\},\{x_{6},x_{7},x_{8}\}\};\\
\mathscr{C}_{2}&=&\{\{x_{1},x_{3},x_{4}\},\{x_{2},x_{3}\},\{x_{4},x_{5}\},
\{x_{5},x_{6}\},\{x_{6}\},\{x_{7},x_{8}\}\};\\
\mathscr{C}_{3}&=&\{\{x_{1}\},\{x_{1},x_{2},x_{3}\},\{x_{2},x_{3}\},
\{x_{3},x_{4},x_{5},x_{6}\},\{x_{5},x_{7},x_{8}\}\};\\
\mathscr{C}_{4}&=&\{\{x_{1},x_{2},x_{4}\},\{x_{2},x_{3}\},
\{x_{4},x_{5},x_{6}\},\{x_{6}\},\{x_{7},x_{8}\}\};\\
\mathscr{C}_{5}&=&\{\{x_{1},x_{2},x_{3}\},\{x_{4}\},
\{x_{5},x_{6}\},\{x_{5},x_{6},x_{8}\},\{x_{4},x_{7},x_{8}\}\};\\
\mathscr{C}_{6}&=&\{\{x_{1},x_{4},x_{5}\},\{x_{2}\},\{x_{3},x_{4},x_{6}\},\{x_{3},x_{5},x_{7}\},
\{x_{7},x_{8}\}\}.
\end{eqnarray*}
By Definition 3.1, we see that $(U,\Delta^{+}, D)$ is a dynamic information system of $(U,\Delta, D)$. Especially, $(U,\Delta, D)$ and $(U,\Delta^{+}, D)$ are consistent covering information systems.
\end{example}

Suppose $(U,\Delta^{+}, D)$ and $(U,\Delta, D)$ are covering information systems, where $U=\{x_{1},x_{2},...,x_{n}\}$, $\Delta=\{\mathscr{C}_{1},\mathscr{C}_{2},...,\mathscr{C}_{m}\}$, and $\Delta^{+}=\{\mathscr{C}_{1},\mathscr{C}_{2},...,\mathscr{C}_{m},\mathscr{C}_{m+1}\}$,
$\mathscr{A}_{\Delta}=\{C_{k}\in \cup \Delta\mid \exists D_{j}\in U/D, \text{ s.t. } C_{k}\subseteq D_{j}\}$, $\mathscr{A}_{\Delta^{+}}=\{C_{k}\in \cup \Delta^{+}\mid \exists D_{j}\in U/D, \text{ s.t. } C_{k}\subseteq D_{j}\}$, $\mathscr{A}_{\mathscr{C}_{m+1}}=\{C_{k}\in \mathscr{C}_{m+1}\mid \exists D_{j}\in U/D, \text{ s.t. } C_{k}\subseteq D_{j}\}$, $r(x)=\{\mathscr{C}\in \Delta\mid \exists C_{k}\in \mathscr{A}_{\Delta}, \text{ s.t. } x\in C_{k}\in \mathscr{C}\},$ and  $r^{+}(x)=\{\mathscr{C}\in \Delta^{+}\mid \exists C_{k}\in \mathscr{A}_{\Delta^{+}}, \text{ s.t. } x\in C_{k}\in \mathscr{C}\}.$

\begin{theorem}
Let $(U,\Delta, D)$ and $(U,\Delta^{+}, D)$ be covering information systems, where $U=\{x_{1},x_{2},...,x_{n}\}$, $\Delta=\{\mathscr{C}_{1},\mathscr{C}_{2},...,\mathscr{C}_{m}\}$, and $\Delta^{+}=\{\mathscr{C}_{1},\mathscr{C}_{2},...,\mathscr{C}_{m},\mathscr{C}_{m+1}\}$. Then we have
\makeatother $$r^{+}(x)=\left\{
\begin{array}{ccc}
r(x)\cup \{\mathscr{C}_{m+1}\},&{\rm if}& x\in \cup\mathscr{A}_{\mathscr{C}_{m+1}};\\
r(x),&{\rm }& otherwise.
\end{array}
\right. $$
\end{theorem}

\noindent\textbf{Proof:} By Definition 2.6, we have $r(x)=\{\mathscr{C}\in \Delta\mid \exists C_{k}\in \mathscr{A}_{\Delta}, \text{ s.t. } x\in C_{k}\in \mathscr{C}\},$ and  $r^{+}(x)=\{\mathscr{C}\in \Delta^{+}\mid \exists C_{k}\in \mathscr{A}_{\Delta^{+}}, \text{ s.t. } x\in C_{k}\in \mathscr{C}\}.$ Since $\Delta^{+}=\{\mathscr{C}_{1},\mathscr{C}_{2},...,\mathscr{C}_{m},\mathscr{C}_{m+1}\}$, it follows that $r^{+}(x)=\{\mathscr{C}\in \Delta\mid \exists C_{k}\in \mathscr{A}_{\Delta}, \text{ s.t. } x\in C_{k}\in \mathscr{C}\}\cup \{\mathscr{C}_{m+1}\mid \exists C\in \mathscr{C}_{m+1}, \text{ s.t. } x\in C_{k}\in \mathscr{C}_{m+1}\}$ for $x\in U$. For simplicity, we denote $\cup\mathscr{A}_{\mathscr{C}_{m+1}}=\{C\mid C\in \mathscr{C}_{m+1}, \exists X\in U/D, \text{ s.t. } C \subseteq D\}$. So we have $r^{+}(x)=r(x)\cup \{\mathscr{C}_{m+1}\}$ and $r^{+}(y)=r(y)$ for $x\in \cup\mathscr{A}_{\mathscr{C}_{m+1}}$ and $y\notin \cup\mathscr{A}_{\mathscr{C}_{m+1}}$, respectively. Therefore, we have
\makeatother $$r^{+}(x)=\left\{
\begin{array}{ccc}
r(x)\cup \{\mathscr{C}_{m+1}\},&{\rm if}& x\in \cup\mathscr{A}_{\mathscr{C}_{m+1}};\\
r(x),&{\rm }& otherwise.
\end{array}
\right. \Box$$

Theorem 3.3 illustrates the relationship between $r(x)$ of $(U,\Delta, D)$ and $r^{+}(x)$ of $(U,\Delta^{+}, D)$, which reduces the time complexity of computing related family $R(U,\Delta^{+}, D)$. Especially, we only need to compute
$\mathscr{A}_{\mathscr{C}_{m+1}}$ for attribute reduction of $(U,\Delta^{+}, D)$, and we get $r^{+}(x)=r(x)$ and  $r^{+}(x)=r(x)\cup \{\mathscr{C}_{m+1}\}$ when $\cup\mathscr{A}_{\mathscr{C}_{m+1}}=\emptyset$ and $\cup\mathscr{A}_{\mathscr{C}_{m+1}}=U$, respectively, for $x\in U.$

\begin{theorem}
Let $(U,\Delta^{+}, D)$ and $(U,\Delta, D)$ be covering information systems, where $U=\{x_{1},x_{2},...,x_{n}\}$, $\Delta=\{\mathscr{C}_{1},\mathscr{C}_{2},...,\mathscr{C}_{m}\}$, $\Delta^{+}=\{\mathscr{C}_{1},\mathscr{C}_{2},...,\mathscr{C}_{m},\mathscr{C}_{m+1}\}$, $\bigtriangleup g(U,\Delta,D)$ $=\bigvee^{k'}_{i=1}\{\bigwedge \Delta'_{i}\mid \Delta'_{i}\subseteq\Delta^{+}\}$ is the reduced disjunctive form of $\bigtriangleup f(U,\Delta^{+},D)$, where $\bigtriangleup f(U,\Delta,D)=(\{\mathscr{C}_{m+1}\})\bigwedge (\bigwedge_{x\notin \cup\mathscr{A}_{\mathscr{C}_{m+1}}}\bigvee r(x))$, and $\bigtriangleup\mathscr{R}(U,\Delta,D)=\{\Delta'_{j}\mid \overline{\exists}\Delta_{i}\in \mathscr{R}(U,\Delta,D), \text{ s.t. } \Delta_{i}\subset\Delta'_{j},1\leq j\leq k'\}$. Then we have $\mathscr{R}(U,\Delta^{+},D)=\mathscr{R}(U,\Delta,D)\cup (\bigtriangleup\mathscr{R}(U,\Delta,D))$.
\end{theorem}

\noindent\textbf{Proof:} On one hand, taking $\Delta_{i}\in \mathscr{R}(U,\Delta,D)$, by Definition 2.5, we have $POS_{\cup\Delta}(D)= POS_{\cup\Delta_{i}}(D)=U$ and $POS_{\cup\Delta_{i}}(D)\neq POS_{\cup\Delta_{i}-\{\mathscr{C}_{i}\}}(D)$ for $\mathscr{C}_{i}\in \Delta_{i}$. We also get $POS_{\cup\Delta^{+}}(D)= POS_{\cup\Delta_{i}}(D)=U$ and $POS_{\cup\Delta_{i}}(D)\neq POS_{\cup\Delta_{i}-\{\mathscr{C}_{i}\}}(D)$ for $\mathscr{C}_{i}\in \Delta_{i}$.
So $\Delta_{i}\in\mathscr{R}(U,\Delta^{+},D)$. Thus, we obtain $\mathscr{R}(U,\Delta,D)\subseteq\mathscr{R}(U,\Delta^{+},D)$. Furthermore, taking $\Delta'_{j}\in \bigtriangleup\mathscr{R}(U,\Delta,D),$ it implies that $POS_{\cup\Delta}(D)= POS_{\cup\Delta'_{j}}(D)=U$ and $POS_{\cup\Delta'_{j}}(D)\neq POS_{\cup\Delta'_{j}-\{\mathscr{C}_{i}\}}(D)$ for $\mathscr{C}_{i}\in \Delta'_{j}$. It follows that $\Delta'_{j}\in \mathscr{R}(U,\Delta^{+},D)$. So we have
$\mathscr{R}(U,\Delta,D)\cup (\bigtriangleup\mathscr{R}(U,\Delta,D))\subseteq\mathscr{R}(U,\Delta^{+},D)$.

On the other hand, we have $\mathscr{R}(U,\Delta^{+},D)=\mathscr{R}_{1}(U,\Delta^{+},D)\cup \mathscr{R}_{2}(U,\Delta^{+},D)$, where $\mathscr{R}_{1}(U,\Delta^{+},D)=\{\Delta_{i}\mid \mathscr{C}_{m+1}\notin \Delta_{i},\Delta_{i}\in \mathscr{R}(U,\Delta^{+},D)\}$ and $\mathscr{R}_{2}(U,\Delta^{+},D)=\{\Delta_{i}\mid \mathscr{C}_{m+1}\in \Delta_{i},\Delta_{i}\in \mathscr{R}(U,\Delta^{+},D)\}$. Obviously, we have $\bigtriangleup\mathscr{R}(U,\Delta,D)\subseteq \mathscr{R}_{2}(U,\Delta^{+},D).$ To prove $\mathscr{R}_{2}(U,\Delta^{+},D)\subseteq \bigtriangleup\mathscr{R}(U,\Delta,D)$, we only need to prove $\mathscr{R}_{2}(U,\Delta^{+},D)\backslash (\bigtriangleup\mathscr{R}(U,\Delta,D))=\emptyset$.
Suppose we have $\Delta'=\{\mathscr{C}_{1'},\mathscr{C}_{2'},...,\mathscr{C}_{k'},\mathscr{C}_{m+1}\}\in \mathscr{R}_{2}(U,\Delta^{+},D)\backslash \bigtriangleup\mathscr{R}(U,\Delta,D),$ there exists $x\in U$ such that $\mathscr{C}_{i'}\in r^{+}(x)\in Md_{\cup R(U,\Delta^{+},D)}(x).$ If $\mathscr{C}_{m+1}\in r^{+}(x)$, then $\mathscr{C}_{i'}$ is superfluous relative to $D$. It implies that $\mathscr{C}_{m+1}\notin r^{+}(x)$. It follows that $\Delta'\in \bigtriangleup\mathscr{R}(U,\Delta,D)$, which is contradicted. So $\mathscr{R}_{2}(U,\Delta^{+},D)\backslash (\bigtriangleup\mathscr{R}(U,\Delta,D))=\emptyset$. Thus $\bigtriangleup\mathscr{R}(U,\Delta,D)=\mathscr{R}_{2}(U,\Delta^{+},D).$

Therefore, we have $\mathscr{R}(U,\Delta^{+},D)=\mathscr{R}(U,\Delta,D)\cup (\bigtriangleup\mathscr{R}(U,\Delta,D))$.
$\Box$

Theorem 3.4 illustrates the relationship between $\mathscr{R}(U,\Delta^{+},D)$ of $(U,\Delta^{+},$ $D)$ and $\mathscr{R}(U,\Delta,D)$ of $(U,\Delta, D)$, which reduces the time complexities of computing reducts of $(U,\Delta^{+}, D)$. Especially, we only need to construct $\bigtriangleup\mathscr{R}(U,\Delta,D)$ for attribute reduction of $(U,\Delta^{+}, D)$.

We provide an incremental algorithm of computing $\mathscr{R}(U,\Delta^{+},D)$ for dynamic covering information system $(U,\Delta^{+},D)$ as follows.

\begin{algorithm}(Incremental Algorithm of Computing $\mathscr{R}(U,\Delta^{+},D)$ for Consistent Covering Information System $(U,\Delta^{+},D)$)(IACAIS)

Step 1: Input $(U,\Delta^{+}, D)$;

Step 2: Construct $POS_{\cup\Delta^{+}}(D)=POS_{\cup\Delta}(D)$;

Step 3: Compute $R(U,\Delta^{+},D)=\{r^{+}(x)\mid x_{i}\in U,r^{+}(x)\neq \emptyset\}$, where \begin{eqnarray*}
r^{+}(x)=\left\{
\begin{array}{ccc}
r(x)\cup \{\mathscr{C}_{m+1}\},&{\rm if}& x\in \cup\mathscr{A}_{\mathscr{C}_{m+1}};\\
r(x),&{\rm }& otherwise.
\end{array}
\right.
\end{eqnarray*}

Step 4: Construct $\triangle f(U,\Delta,D)=(\{\mathscr{C}_{m+1}\})\bigwedge (\bigwedge_{x\notin \cup\mathscr{A}_{\mathscr{C}_{m+1}}}\bigvee r(x))$;

Step 5: Compute $\bigtriangleup\mathscr{R}(U,\Delta,D)=\{\Delta'_{j}\mid \overline{\exists}\Delta_{i}\in \mathscr{R}(U,\Delta,D), \text{ s.t. } \Delta_{i}\subset\Delta'_{j},1\leq j\leq k'\}$;

Step 6: Output $\mathscr{R}(U,\Delta^{+},D)=\mathscr{R}(U,\Delta,D)\cup (\triangle\mathscr{R}(U,\Delta,D))$.
\end{algorithm}

The time complexity of Step 3 is $[|U|\ast|\mathscr{C}_{m+1}|,|U|\ast|\mathscr{C}_{m+1}|\ast |U/D|]$; the time complexity of Steps 4 and 5 is $[|U|-|\cup\mathscr{A}_{\mathscr{C}_{m+1}}|,|U|\ast(|\Delta|+1)]$. Therefore, the
time complexity of the incremental algorithm is lower than that of
the non-incremental algorithm.

\begin{example}(Continuation from Example 3.2)
By Definition 2.6, we first have $
r(x_{1})=\{\mathscr{C}_{1},\mathscr{C}_{3},\mathscr{C}_{5}\},
r(x_{2})$ $=\{\mathscr{C}_{1},\mathscr{C}_{2},\mathscr{C}_{3},\mathscr{C}_{4},\mathscr{C}_{5}\},
r(x_{3})=\{\mathscr{C}_{1},\mathscr{C}_{2},\mathscr{C}_{3},\mathscr{C}_{4},\mathscr{C}_{5}\},
r(x_{4})=\{\mathscr{C}_{1},\mathscr{C}_{2},\mathscr{C}_{4},\mathscr{C}_{5}\},
r(x_{5})=\{\mathscr{C}_{1},\mathscr{C}_{2},\mathscr{C}_{4},\mathscr{C}_{5}\},
r(x_{6})$ $=\{\mathscr{C}_{1},\mathscr{C}_{2},\mathscr{C}_{4},\mathscr{C}_{5}\},
r(x_{7})=\{\mathscr{C}_{2},\mathscr{C}_{4}\},$ and $
r(x_{8})=\{\mathscr{C}_{2},\mathscr{C}_{4}\}.$
Thus, we get
$R(U,\Delta,D)=\{\{\mathscr{C}_{1},\mathscr{C}_{3},\mathscr{C}_{5}\},$ $
\{\mathscr{C}_{1},$ $\mathscr{C}_{2},\mathscr{C}_{3},\mathscr{C}_{4},\mathscr{C}_{5}\},
\{\mathscr{C}_{1},\mathscr{C}_{2},\mathscr{C}_{4},\mathscr{C}_{5}\}
\{\mathscr{C}_{2},\mathscr{C}_{4}\}\}.$
After that, by Definition 2.7, we obtain
\begin{eqnarray*}
f(U,\Delta,D)&=&\bigwedge\{\bigvee r(x)\mid r(x)\in R(U,\Delta,D)\}\\
&=&(\mathscr{C}_{1}\vee\mathscr{C}_{3}\vee\mathscr{C}_{5})\wedge(\mathscr{C}_{1}\vee\mathscr{C}_{2}\vee\mathscr{C}_{3}\vee\mathscr{C}_{4}\vee\mathscr{C}_{5})\wedge
(\mathscr{C}_{1}\vee\mathscr{C}_{2}\vee\mathscr{C}_{4}\vee\mathscr{C}_{5})
\wedge(\mathscr{C}_{2}\vee\mathscr{C}_{4})\\
&=&(\mathscr{C}_{1}\vee\mathscr{C}_{3}\vee\mathscr{C}_{5})\wedge (\mathscr{C}_{2}\vee\mathscr{C}_{4})\\
&=&(\mathscr{C}_{1}\wedge\mathscr{C}_{2})\vee (\mathscr{C}_{1}\wedge\mathscr{C}_{4})\vee (\mathscr{C}_{2}\wedge\mathscr{C}_{3})\vee(\mathscr{C}_{3}\wedge\mathscr{C}_{4})\vee
(\mathscr{C}_{2}\wedge\mathscr{C}_{5})\vee(\mathscr{C}_{4}\wedge\mathscr{C}_{5}).
\end{eqnarray*}
So we have $\mathscr{R}(\Delta,U,D)=\{\{\mathscr{C}_{1},\mathscr{C}_{2}\}, \{\mathscr{C}_{1},\mathscr{C}_{4}\}, \{\mathscr{C}_{2},\mathscr{C}_{3}\},\{\mathscr{C}_{3},\mathscr{C}_{4}\},
\{\mathscr{C}_{2},\mathscr{C}_{5}\},\{\mathscr{C}_{4},\mathscr{C}_{5}\}\}.$

Secondly, by Definition 2.6, we have
$r^{+}(x_{1})=r(x_{1}),
r^{+}(x_{2})=r(x_{2})\cup\{\mathscr{C}_{6}\},
r^{+}(x_{3})=r(x_{3}),
r^{+}(x_{4})=r(x_{4}),
r^{+}(x_{5})=r(x_{5}),
r^{+}(x_{6})=r(x_{6}),
r^{+}(x_{7})=r(x_{7})\cup \{\mathscr{C}_{6}\},$ and $
r^{+}(x_{8})=r(x_{8})\cup \{\mathscr{C}_{6}\}.$ By Definition 2.6, we get
$R(U,\Delta^{+},D)=\{\{\mathscr{C}_{1},\mathscr{C}_{3},\mathscr{C}_{5}\},
\{\mathscr{C}_{1},$ $\mathscr{C}_{2},\mathscr{C}_{3},\mathscr{C}_{4},\mathscr{C}_{5}\},\{\mathscr{C}_{1},\mathscr{C}_{2},\mathscr{C}_{3},\mathscr{C}_{4},\mathscr{C}_{5},\mathscr{C}_{6}\},\{\mathscr{C}_{1},\mathscr{C}_{2},\mathscr{C}_{4},\mathscr{C}_{5}\},
\{\mathscr{C}_{2},\mathscr{C}_{4},\mathscr{C}_{6}\}\}.$
By Definition 2.7, we obtain
\begin{eqnarray*}
f(U,\Delta^{+},D)&=&\bigwedge\{\bigvee r^{+}(x)\mid r^{+}(x)\in R(U,\Delta^{+},D)\}\\
&=&(\mathscr{C}_{1}\vee\mathscr{C}_{3}\vee\mathscr{C}_{5})\wedge(\mathscr{C}_{1}\vee\mathscr{C}_{2}
\vee\mathscr{C}_{3}\vee\mathscr{C}_{4}\vee\mathscr{C}_{5}\vee\mathscr{C}_{6})\wedge
(\mathscr{C}_{1}\vee\mathscr{C}_{2}\vee\mathscr{C}_{3}\vee\mathscr{C}_{4}
\vee\mathscr{C}_{5})\wedge
(\mathscr{C}_{1}\vee\mathscr{C}_{2}\\&&\vee\mathscr{C}_{4}\vee\mathscr{C}_{5})
\wedge(\mathscr{C}_{2}\vee\mathscr{C}_{4}\vee\mathscr{C}_{6})\\
&=&(\mathscr{C}_{1}\vee\mathscr{C}_{3}\vee\mathscr{C}_{5})\wedge (\mathscr{C}_{1}\vee\mathscr{C}_{2}\vee\mathscr{C}_{4}\vee\mathscr{C}_{5})\wedge (\mathscr{C}_{2}\vee\mathscr{C}_{4}\vee\mathscr{C}_{6})\\
&=&(\mathscr{C}_{1}\wedge\mathscr{C}_{2})\vee (\mathscr{C}_{1}\wedge\mathscr{C}_{4})\vee (\mathscr{C}_{1}\wedge\mathscr{C}_{6})\vee (\mathscr{C}_{2}\wedge\mathscr{C}_{3})\vee
(\mathscr{C}_{2}\wedge\mathscr{C}_{5})\vee
(\mathscr{C}_{3}\wedge\mathscr{C}_{4})\vee
(\mathscr{C}_{4}\wedge\mathscr{C}_{5})\vee\\&&
(\mathscr{C}_{5}\wedge\mathscr{C}_{6}).
\end{eqnarray*}
So we have $\mathscr{R}(\Delta^{+},D,U)=\{\{\mathscr{C}_{1},\mathscr{C}_{2}\}, \{\mathscr{C}_{1},\mathscr{C}_{4}\}, \{\mathscr{C}_{1},\mathscr{C}_{6}\}, \{\mathscr{C}_{2},\mathscr{C}_{3}\},
\{\mathscr{C}_{2},\mathscr{C}_{5}\},
\{\mathscr{C}_{3},\mathscr{C}_{4}\},
\{\mathscr{C}_{4},\mathscr{C}_{5}\},
\{\mathscr{C}_{5},\mathscr{C}_{6}\}\}.$

Thirdly, by Theorem 3.5, we get
\begin{eqnarray*}
\triangle  f(U,\Delta^{+},D)&=&\mathscr{C}_{6}\wedge(\mathscr{C}_{1}\vee\mathscr{C}_{3}\vee\mathscr{C}_{5})\wedge(\mathscr{C}_{1}\vee\mathscr{C}_{2}\vee\mathscr{C}_{3}\vee
\mathscr{C}_{4}\vee\mathscr{C}_{5})\wedge
(\mathscr{C}_{1}\vee\mathscr{C}_{2}\vee\mathscr{C}_{4}\vee\mathscr{C}_{5})\\
&=&\mathscr{C}_{6}\wedge(\mathscr{C}_{1}\vee\mathscr{C}_{3}\vee\mathscr{C}_{5})\wedge (\mathscr{C}_{1}\vee\mathscr{C}_{2}\vee\mathscr{C}_{4}\vee\mathscr{C}_{5})\\
&=&(\mathscr{C}_{1}\wedge\mathscr{C}_{6})\vee (\mathscr{C}_{5}\wedge\mathscr{C}_{6}).
\end{eqnarray*}

Therefore, we have $\mathscr{R}(\Delta^{+},D,U)=\{\{\mathscr{C}_{1},\mathscr{C}_{2}\}, \{\mathscr{C}_{1},\mathscr{C}_{4}\},  \{\mathscr{C}_{2},\mathscr{C}_{3}\},
\{\mathscr{C}_{2},\mathscr{C}_{5}\},
\{\mathscr{C}_{3},\mathscr{C}_{4}\},
\{\mathscr{C}_{4},\mathscr{C}_{5}\},\{\mathscr{C}_{1},\mathscr{C}_{6}\},$ $
\{\mathscr{C}_{5},\mathscr{C}_{6}\}\}.$
\end{example}

Example 3.6 illustrates how to compute attribute reducts of $(\Delta^{+},D,U)$ by Algorithm 2.8; Example 3.6 also illustrates how to compute attribute reducts of $(\Delta^{+},D,U)$ by Algorithm 3.5. We see that the incremental algorithm is more effective than the non-incremental algorithm for attribute reduction of dynamic covering decision information systems.

In practical situations, there are a lot of dynamic covering information systems caused by deleting attributes, and we also study attribute reduction of dynamic covering information systems when deleting attributes as follows.

\begin{definition}
Let $(U,\Delta, D)$ and $(U,\Delta^{-}, D)$ be covering information systems, where $U=\{x_{1},x_{2},...,x_{n}\}$, $\Delta=\{\mathscr{C}_{1},\mathscr{C}_{2},...,\mathscr{C}_{m}\}$, and $\Delta^{-}=\{\mathscr{C}_{1},\mathscr{C}_{2},...,\mathscr{C}_{m-1}\}$. Then $(U,\Delta^{-}, D)$ is called a dynamic covering information system of $(U,\Delta, D)$.
\end{definition}

\noindent\textbf{Remark:} We take $(U,\Delta, D)$ as a consistent covering information system in Definition 3.7. We also notice that the dynamic covering information system $(U,\Delta^{-}, D)$ is consistent or inconsistent when deleting $\mathscr{C}_{m}$ from $(U,\Delta,D)$.

\begin{example}
Let $(U,\Delta, D)$ and $(U,\Delta^{-}, D)$ be covering information systems, where $U=\{x_{1},x_{2},...,x_{8}\}$, $\Delta=\{\mathscr{C}_{1},
\mathscr{C}_{2},\mathscr{C}_{3},\mathscr{C}_{4},\mathscr{C}_{5}\}$, $\Delta^{-}=\{\mathscr{C}_{1},
\mathscr{C}_{2},\mathscr{C}_{3},\mathscr{C}_{4}\}$, and $U/D=\{\{x_{1},x_{2},x_{3}\},\{x_{4},x_{5},x_{6}\},\{x_{7},x_{8}\}\}$, where
\begin{eqnarray*}
\mathscr{C}_{1}&=&\{\{x_{1},x_{2}\},\{x_{2},x_{3},x_{4}\},\{x_{3}\},
\{x_{4}\},\{x_{5},x_{6}\},\{x_{6},x_{7},x_{8}\}\};\\
\mathscr{C}_{2}&=&\{\{x_{1},x_{3},x_{4}\},\{x_{2},x_{3}\},\{x_{4},x_{5}\},
\{x_{5},x_{6}\},\{x_{6}\},\{x_{7},x_{8}\}\};\\
\mathscr{C}_{3}&=&\{\{x_{1}\},\{x_{1},x_{2},x_{3}\},\{x_{2},x_{3}\},
\{x_{3},x_{4},x_{5},x_{6}\},\{x_{5},x_{7},x_{8}\}\};\\
\mathscr{C}_{4}&=&\{\{x_{1},x_{2},x_{4}\},\{x_{2},x_{3}\},
\{x_{4},x_{5},x_{6}\},\{x_{6}\},\{x_{7},x_{8}\}\};\\
\mathscr{C}_{5}&=&\{\{x_{1},x_{2},x_{3}\},\{x_{4}\},
\{x_{5},x_{6}\},\{x_{5},x_{6},x_{8}\},\{x_{4},x_{7},x_{8}\}\}.
\end{eqnarray*}
By Definition 3.7, we see that $(U,\Delta^{-}, D)$ is a dynamic covering information system of $(U,\Delta, D)$. Especially, $(U,\Delta, D)$ and $(U,\Delta^{-}, D)$ are consistent covering information systems.
\end{example}

Suppose $(U,\Delta, D)$ and $(U,\Delta^{-}, D)$ are covering information systems, where $U=\{x_{1},x_{2},...,x_{n}\}$, $\Delta=\{\mathscr{C}_{1},\mathscr{C}_{2},...,\mathscr{C}_{m}\}$, and $\Delta^{-}=\{\mathscr{C}_{1},\mathscr{C}_{2},...,\mathscr{C}_{m-1}\}$,
$\mathscr{A}_{\Delta}=\{C_{k}\in \cup \Delta\mid \exists D_{j}\in U/D, \text{ s.t. } C_{k}\subseteq D_{j}\}$, $r(x)=\{\mathscr{C}\in \Delta\mid \exists C_{k}\in \mathscr{A}_{\Delta}, \text{ s.t. } x\in C_{k}\in \mathscr{C}\},$ and  $r^{-}(x)=\{\mathscr{C}\in \Delta^{-}\mid \exists C_{k}\in \mathscr{A}_{\Delta^{-}}, \text{ s.t. } x\in C_{k}\in \mathscr{C}\}.$

\begin{theorem}
Let $(U,\Delta, D)$ and $(U,\Delta^{-}, D)$ be covering information systems, where $U=\{x_{1},x_{2},...,x_{n}\}$, $\Delta=\{\mathscr{C}_{1},\mathscr{C}_{2},...,\mathscr{C}_{m}\}$, and $\Delta^{-}=\{\mathscr{C}_{1},\mathscr{C}_{2},...,\mathscr{C}_{m-1}\}$. Then
we have $\makeatother r^{-}(x)=r(x)\backslash\{\mathscr{C}_{m}\}.$
\end{theorem}

\noindent\textbf{Proof:} The proof is straightforward by Definitions 2.6 and 3.7.$\Box$

Theorem 3.9 illustrates the relationship between $r(x)$ of $(U,\Delta, D)$ and $r^{-}(x)$ of $(U,\Delta^{-}, D)$, which reduces the time complexities of computing related family $R(U,\Delta^{-}, D)$. In other words, we obtain the related family $R(U,\Delta^{-}, D)$ with lower time complexity.

\begin{theorem}
Let $(U,\Delta, D)$ and $(U,\Delta^{-}, D)$ be covering information systems, where $U=\{x_{1},x_{2},...,x_{n}\}$, $\Delta=\{\mathscr{C}_{1},\mathscr{C}_{2},...,\mathscr{C}_{m}\}$, and $\Delta^{-}=\{\mathscr{C}_{1},\mathscr{C}_{2},...,\mathscr{C}_{m-1}\}$. If $(U,\Delta^{-}, D)$ is a dynamic consistent covering information system, then
we have $\mathscr{R}(U,\Delta^{-},D)=\{\Delta_{i}\mid \mathscr{C}_{m}\notin\Delta_{i}\in \mathscr{R}(U,\Delta,D)\}.$
\end{theorem}

\noindent\textbf{Proof:} The proof is straightforward by Definition 3.7 and Theorem 3.9.$\Box$

Theorem 3.10 illustrates the relationship between $\mathscr{R}(U,\Delta^{-},D)$ of  $(U,\Delta, D)$ and $\mathscr{R}(U,\Delta,D)$ of $(U,\Delta^{-},$ $D)$, and
we have $\mathscr{R}(U,\Delta^{-},D)=\{\Delta_{i}\mid \mathscr{C}_{m}\notin\Delta_{i}\in \mathscr{R}(U,\Delta,D)\}$ when $POS_{\cup\Delta^{-}}(D)=POS_{\cup\Delta}(D)$, which reduces the time complexities of computing attribute reducts of dynamic covering information systems.

\begin{theorem}
Let $(U,\Delta, D)$ and $(U,\Delta^{-}, D)$ be covering information systems, where $U=\{x_{1},x_{2},$ $...,x_{n}\}$, $\Delta=\{\mathscr{C}_{1},\mathscr{C}_{2},...,\mathscr{C}_{m}\}$, and $\Delta^{-}=\{\mathscr{C}_{1},\mathscr{C}_{2},...,\mathscr{C}_{m-1}\}$. If $(U,\Delta^{-}, D)$ is a dynamic inconsistent covering information system, then
we have $\mathscr{R}(U,\Delta^{-},D)=\{\Delta_{i}\backslash \{\mathscr{C}_{m}\}\mid \Delta_{i}\in \mathscr{R}(U,\Delta,D)\}.$
\end{theorem}

\noindent\textbf{Proof:} The proof is straightforward by Definition 3.7 and Theorem 3.9.$\Box$

Theorem 3.11 illustrates the relationship between $\mathscr{R}(U,\Delta^{-},D)$ of $(U,\Delta^{-},$ $D)$  and $\mathscr{R}(U,\Delta,D)$ of $(U,\Delta, D)$, and we have $\mathscr{R}(U,\Delta^{-},D)=\{\Delta_{i}\backslash \{\mathscr{C}_{m}\}\mid \Delta_{i}\in \mathscr{R}(U,\Delta,D)\}$ when $POS_{\cup\Delta^{-}}(D)\neq POS_{\cup\Delta}(D)$, which reduces time complexities of computing attribute reducts of dynamic covering information systems.

We provide an incremental algorithm of computing $\mathscr{R}(U,\Delta^{-},D)$ for dynamic covering information system $(U,\Delta^{-},D)$ as follows.

\begin{algorithm}(Incremental Algorithm of Computing $\mathscr{R}(U,\Delta^{-},D)$ for Covering Information System $(U,\Delta^{-},$ $D)$)(IADCIS)

Step 1: Input $(U,\Delta^{-}, D)$;

Step 2: Construct $POS_{\cup\Delta^{-}}(D)$;

Step 3: Compute $\mathscr{R}(U,\Delta^{-},D)=\{\Delta_{i}\mid \mathscr{C}_{m}\notin\Delta_{i}\in \mathscr{R}(U,\Delta,D)\}$ when $POS_{\cup\Delta^{-}}(D)=POS_{\cup\Delta}(D)$;

Step 4: Construct $\mathscr{R}(U,\Delta^{-},D)=\{\Delta_{i}\backslash \{\mathscr{C}_{m}\}\mid \Delta_{i}\in \mathscr{R}(U,\Delta,D)\}$ when $POS_{\cup\Delta^{-}}(D)\neq POS_{\cup\Delta}(D)$;

Step 5: Output $\mathscr{R}(\Delta^{-},U,D)$.
\end{algorithm}

The time complexity of Step 2 is $[|U|\ast(\sum_{\mathscr{C}\in\Delta^{-}}|\mathscr{C}|),|U|\ast(\sum_{\mathscr{C}\in\Delta^{-}}|\mathscr{C}|)\ast |U/D|]$; the time complexity of Step 3 is $[|U|^{2},|U|^{2}\ast(\sum_{\mathscr{C}\in\Delta}|\mathscr{C}|)\ast |U/D|]$; the time complexity of Steps 3 and 4 is $|\mathscr{R}(U,\Delta,D)|$. Therefore, the
time complexity of the non-incremental algorithm is very high. Therefore, the
time complexity of the incremental algorithm is lower than that of
the non-incremental algorithm.

\begin{example}(Continuation from Example 3.2)
By Definition 3.1, we first have $
r^{-}(x_{1})=\{\mathscr{C}_{1},\mathscr{C}_{3}\},
r^{-}(x_{2})$ $=\{\mathscr{C}_{1},\mathscr{C}_{2},\mathscr{C}_{3},\mathscr{C}_{4}\},
r^{-}(x_{3})=\{\mathscr{C}_{1},\mathscr{C}_{2},\mathscr{C}_{3},\mathscr{C}_{4}\},
r^{-}(x_{4})=\{\mathscr{C}_{1},\mathscr{C}_{2},\mathscr{C}_{4}\},
r^{-}(x_{5})=\{\mathscr{C}_{1},\mathscr{C}_{2},\mathscr{C}_{4}\},
r^{-}(x_{6})=\{\mathscr{C}_{1},$ $\mathscr{C}_{2},\mathscr{C}_{4}\},$ $
r^{-}(x_{7})=\{\mathscr{C}_{2},\mathscr{C}_{4}\}, $ and $
r^{-}(x_{8})=\{\mathscr{C}_{2},\mathscr{C}_{4}\}.$
So we have
\begin{eqnarray*}
f(U,\Delta^{-},D)&=&\bigwedge\{\bigvee r^{-}(x)\mid r^{-}(x)\in R(U,\Delta,D)\}\\
&=&(\mathscr{C}_{1}\vee\mathscr{C}_{3})\wedge(\mathscr{C}_{2}\vee\mathscr{C}_{4})\\
&=&(\mathscr{C}_{1}\wedge\mathscr{C}_{2})\vee (\mathscr{C}_{1}\wedge\mathscr{C}_{4})\vee (\mathscr{C}_{2}\wedge\mathscr{C}_{3})\vee (\mathscr{C}_{3}\wedge\mathscr{C}_{4}).
\end{eqnarray*}
Thus, we have $\mathscr{R}(U,\Delta^{-},D)=\{\{\mathscr{C}_{1},\mathscr{C}_{2}\}, \{\mathscr{C}_{1},\mathscr{C}_{3}\}\}.$

Secondly, by Theorem 3.10,
since $POS_{\cup\Delta^{-}}(D)=POS_{\cup\Delta}(D)=U$,
we have $\mathscr{R}(U,\Delta^{-},D)=\{\{\mathscr{C}_{1},\mathscr{C}_{2}\}, \{\mathscr{C}_{1},$ $\mathscr{C}_{4}\}, \{\mathscr{C}_{2},\mathscr{C}_{3}\},\{\mathscr{C}_{3},\mathscr{C}_{4}\}\}.$
\end{example}

Example 3.13 illustrates how to compute attribute reducts of $(\Delta^{-},D,U)$ by Algorithm 2.8; Example 3.13 also illustrates how to compute attribute reducts of $(\Delta^{-},D,U)$ by Algorithm 3.12. We see that the incremental algorithm is more effective than the non-incremental algorithm for attribute reduction of dynamic covering decision information systems.

\section{Related family-based attribute reduction of inconsistent covering information systems with variations of attributes}

In this section, we study related family-based attribute reduction of inconsistent covering information systems with variations of attributes.

\begin{definition}
Let $(U,\Delta, D)$ and $(U,\Delta^{+}, D)$ be covering information systems, where $U=\{x_{1},x_{2},...,x_{n}\}$, $\Delta=\{\mathscr{C}_{1},
\mathscr{C}_{2},...,\mathscr{C}_{m}\}$, and $\Delta^{+}=\{\mathscr{C}_{1},
\mathscr{C}_{2},...,\mathscr{C}_{m},\mathscr{C}_{m+1}\}$. Then $(U,\Delta^{+}, D)$ is called a dynamic information system of $(U,\Delta, D)$.
\end{definition}

\noindent\textbf{Remark:} We take $(U,\Delta, D)$ as an inconsistent covering information system in Definition 4.1. We also notice that the dynamic covering information system $(U,\Delta^{+}, D)$ is consistent or inconsistent when adding $\mathscr{C}_{m+1}$ into $(U,\Delta,D)$. In practical situations, there are many dynamic covering information systems, and we only discuss inconsistent covering information systems with variations of attributes in this section.

\begin{example}
Let $(U,\Delta, D)$ and $(U,\Delta^{+}, D)$ be covering information systems, where $U=\{x_{1},x_{2},...,x_{8}\}$, $\Delta=\{\mathscr{C}_{1},
\mathscr{C}_{2},\mathscr{C}_{3},\mathscr{C}_{4}\}$, $\Delta^{+}=\{\mathscr{C}_{1},
\mathscr{C}_{2},\mathscr{C}_{3},\mathscr{C}_{4},\mathscr{C}_{5}\}$, and $U/D=\{\{x_{1},x_{2},x_{3}\},\{x_{4},x_{5},x_{6}\},\{x_{7},x_{8}\}\}$, where
\begin{eqnarray*}
\mathscr{C}_{1}&=&\{\{x_{1},x_{2},x_{3},x_{4}\},\{x_{3},x_{6},x_{7}\},
\{x_{4},x_{5}\},\{x_{6}\},\{x_{7},x_{8}\}\};\\
\mathscr{C}_{2}&=&\{\{x_{1}\},\{x_{2},x_{3},x_{4}\},\{x_{4},x_{5}\},
\{x_{4},x_{5},x_{6}\},\{x_{6},x_{7},x_{8}\}\};\\
\mathscr{C}_{3}&=&\{\{x_{1}\},\{x_{1},x_{3},x_{4}\},\{x_{2},x_{3},x_{4},x_{8}\},
\{x_{3},x_{4},x_{5},x_{6},x_{7}\}\};\\
\mathscr{C}_{4}&=&\{\{x_{1},x_{4},x_{5}\},\{x_{2},x_{3},x_{4},x_{5}\},
\{x_{4},x_{5},x_{6},x_{7},x_{8}\}\};\\
\mathscr{C}_{5}&=&\{\{x_{1},x_{5},x_{6}\},\{x_{4},x_{5}\},
\{x_{2},x_{3},x_{4}\},\{x_{5},x_{6},x_{7},x_{8}\}\}.
\end{eqnarray*}
By Definition 4.1, we see that $(U,\Delta^{+}, D)$ is a dynamic information system of $(U,\Delta, D)$. Especially, $(U,\Delta, D)$ and $(U,\Delta^{+}, D)$ are inconsistent covering information systems.
\end{example}

Suppose $(U,\Delta^{+}, D)$ and $(U,\Delta, D)$ are covering information systems, where $U=\{x_{1},x_{2},...,x_{n}\}$, $\Delta=\{\mathscr{C}_{1},
\mathscr{C}_{2},...,\mathscr{C}_{m}\}$, and $\Delta^{+}=\{\mathscr{C}_{1},
\mathscr{C}_{2},...,\mathscr{C}_{m},\mathscr{C}_{m+1}\}$,
$\mathscr{A}_{\Delta}=\{C_{k}\in \cup \Delta\mid \exists D_{j}\in U/D, \text{ s.t. } C_{k}\subseteq D_{j}\}$, $r(x)=\{\mathscr{C}\in \Delta\mid \exists C_{k}\in \mathscr{A}_{\Delta}, \text{ s.t. } x\in C_{k}\in \mathscr{C}\},$ $\mathscr{A}_{\Delta^{+}}=\{C_{k}\in \cup \Delta^{+}\mid \exists D_{j}\in U/D, \text{ s.t. } C_{k}\subseteq D_{j}\}$, $\mathscr{A}_{\mathscr{C}_{m+1}}=\{C_{k}\in \mathscr{C}_{m+1}\mid \exists D_{j}\in U/D, \text{ s.t. } C_{k}\subseteq D_{j}\}$, $r(x)=\{\mathscr{C}\in \Delta\mid \exists C_{k}\in \mathscr{A}_{\Delta}, \text{ s.t. } x\in C_{k}\in \mathscr{C}\},$ and  $r^{+}(x)=\{\mathscr{C}\in \Delta^{+}\mid \exists C_{k}\in \mathscr{A}_{\Delta^{+}}, \text{ s.t. } x\in C_{k}\in \mathscr{C}\}.$

\begin{theorem}
Let $(U,\Delta, D)$ and $(U,\Delta^{+}, D)$ be covering information systems, where $U=\{x_{1},x_{2},...,x_{n}\}$, $\Delta=\{\mathscr{C}_{1},
\mathscr{C}_{2},...,\mathscr{C}_{m}\}$, and $\Delta^{+}=\{\mathscr{C}_{1},
\mathscr{C}_{2},...,\mathscr{C}_{m},\mathscr{C}_{m+1}\}$. Then we have
\makeatother $$r^{+}(x)=\left\{
\begin{array}{ccc}
r(x)\cup \{\mathscr{C}_{m+1}\},&{\rm if}& x\in \cup\mathscr{A}_{\mathscr{C}_{m+1}};\\
r(x),&{\rm }& otherwise.
\end{array}
\right. $$
\end{theorem}

\noindent\textbf{Proof:} The proof is similar to Theorem 3.3.$\Box$

Theorem 4.3 illustrates the relationship between $r(x)$ of $(U,\Delta, D)$ and $r^{+}(x)$ of $(U,\Delta^{+}, D)$, and  reduces the time complexity of computing related family $R(U,\Delta^{+}, D)$. Especially, we only need to compute
$\mathscr{A}_{\mathscr{C}_{m+1}}$ for attribute reduction of $(U,\Delta^{+}, D)$, and we get $r^{+}(x)=r(x)$ and  $r^{+}(x)=r(x)\cup \{\mathscr{C}_{m+1}\}$ when $\cup\mathscr{A}_{\mathscr{C}_{m+1}}=\emptyset$ and $\cup\mathscr{A}_{\mathscr{C}_{m+1}}=U$, respectively, for $x\in U.$

\begin{theorem}
Let $(U,\Delta^{+}, D)$ and $(U,\Delta, D)$ be covering information systems, where $U=\{x_{1},x_{2},...,x_{n}\}$, $\Delta=\{\mathscr{C}_{1},\mathscr{C}_{2},...,\mathscr{C}_{m}\}$, $\Delta^{+}=\{\mathscr{C}_{1},\mathscr{C}_{2},...,\mathscr{C}_{m},\mathscr{C}_{m+1}\}$, $\bigtriangleup g(U,\Delta,D)$ $=\bigvee^{k'}_{i=1}\{\bigwedge \Delta'_{i}\mid \Delta'_{i}\subseteq\Delta^{+}\}$ is the reduced disjunctive form of $\bigtriangleup f(U,\Delta^{+},D)$, where $\bigtriangleup f(U,\Delta,D)=(\{\mathscr{C}_{m+1}\})\bigwedge (\bigwedge_{x\in POS_{\cup\Delta^{+}}(D)\wedge x\notin \cup\mathscr{A}_{\mathscr{C}_{m+1}}}\bigvee r(x))$, and $\bigtriangleup\mathscr{R}(U,\Delta,D)=\{\Delta'_{j}\mid \overline{\exists}\Delta_{i}\in \mathscr{R}(U,\Delta,D), \text{ s.t. } \Delta_{i}\subset\Delta'_{j},1\leq j\leq k'\}$. If $POS_{\cup\Delta^{+}}(D)=POS_{\cup\Delta}(D)$, then  $\mathscr{R}(U,\Delta^{+},D)=\mathscr{R}(U,\Delta,D)\cup (\bigtriangleup\mathscr{R}(U,\Delta,D))$.
\end{theorem}

\noindent\textbf{Proof:} The proof is similar to Theorem 3.4.$\Box$

Theorem 4.4 illustrates the relationship between $\mathscr{R}(U,\Delta^{+},D)$ and $\mathscr{R}(U,\Delta,D)$, and we get $\mathscr{R}(U,\Delta^{+},D)$ $=\mathscr{R}(U,\Delta,D)\cup (\bigtriangleup\mathscr{R}(U,\Delta,D))$ when $POS_{\cup\Delta^{+}}(D)= POS_{\cup\Delta}(D)$, which reduces the time complexities of computing attribute reducts of dynamic covering information systems.

\begin{theorem}
Let $(U,\Delta^{+}, D)$ and $(U,\Delta, D)$ be covering information systems, where $U=\{x_{1},x_{2},...,x_{n}\}$, $\Delta=\{\mathscr{C}_{1},\mathscr{C}_{2},...,\mathscr{C}_{m}\}$, $\Delta^{+}=\{\mathscr{C}_{1},\mathscr{C}_{2},...,\mathscr{C}_{m},\mathscr{C}_{m+1}\}$, $\bigtriangleup g(U,\Delta,D)$ $=\bigvee^{k'}_{i=1}\{\bigwedge \Delta'_{i}\mid \Delta'_{i}\subseteq\Delta^{+}\}$ is the reduced disjunctive form of $\bigtriangleup f(U,\Delta^{+},D)$, where $\bigtriangleup f(U,\Delta,D)=(\{\mathscr{C}_{m+1}\})\bigwedge (\bigwedge_{x\in POS_{\cup\Delta^{+}}(D)\wedge x\notin \cup\mathscr{A}_{\mathscr{C}_{m+1}}}\bigvee r(x))$, and $\bigtriangleup\mathscr{R}(U,\Delta,D)=\{\Delta'_{j}\mid \overline{\exists}\Delta_{i}\in \mathscr{R}(U,\Delta,D), \text{ s.t. } \Delta_{i}\subset\Delta'_{j},1\leq j\leq k'\}$. If $POS_{\cup\Delta^{+}}(D)\neq POS_{\cup\Delta}(D)$, then  $\mathscr{R}(U,\Delta^{+},D)=\{\Delta_{i}\cup \{\mathscr{C}_{m+1}\}\mid \Delta_{i}\in\mathscr{R}(U,\Delta,D)\}$.
\end{theorem}

\noindent\textbf{Proof:} The proof is similar to Theorem 3.4.$\Box$

Theorem 4.5 illustrates the relationship between $\mathscr{R}(U,\Delta^{+},D)$ and $\mathscr{R}(U,\Delta,D)$, and we get $\mathscr{R}(U,\Delta^{+},D)$ $=\{\Delta_{i}\cup \{\mathscr{C}_{m+1}\}|\Delta_{i}\in\mathscr{R}(U,\Delta,D)\}$ when $POS_{\cup\Delta^{+}}(D)\neq POS_{\cup\Delta}(D)$, which reduces the time complexities of computing attribute reducts of dynamic covering information systems.

\begin{proposition}
Let $(U,\Delta, D)$ and $(U,\Delta^{+}, D)$ be covering information systems, where $U=\{x_{1},x_{2},...,x_{n}\}$, $\Delta=\{\mathscr{C}_{1},
\mathscr{C}_{2},...,\mathscr{C}_{m}\}$, and $\Delta^{+}=\{\mathscr{C}_{1},
\mathscr{C}_{2},...,\mathscr{C}_{m},\mathscr{C}_{m+1}\}$. If $POS_{\cup\Delta^{+}}(D)\neq POS_{\cup\Delta}(D)$, there exists $x\in U\backslash POS_{\cup\Delta}(D)$ such that $r^{+}(x)=\{\mathscr{C}_{m+1}\}.$
\end{proposition}

We provide an
incremental algorithm of computing $\mathscr{R}(U,\Delta^{+},D)$ for dynamic covering information system $(U,\Delta^{+},D)$ as follows.

\begin{algorithm}(Incremental Algorithm of Computing $\mathscr{R}(U,\Delta^{+},D)$ for Covering Information System $(U,\Delta^{+},$ $D)$)(IAIAIS)

Step 1: Input $(U,\Delta^{+}, D)$;

Step 2: Construct $POS_{\cup\Delta^{+}}(D)=POS_{\cup\Delta}(D)$;

Step 3: Compute $R(U,\Delta^{+},D)=\{r^{+}(x)\mid x\in U,r^{+}(x)\neq \emptyset\}$, where \begin{eqnarray*}
r^{+}(x)=\left\{
\begin{array}{ccc}
r(x)\cup \{\mathscr{C}_{m+1}\},&{\rm if}& x\in \cup\mathscr{A}_{\mathscr{C}_{m+1}};\\
r(x),&{\rm }& otherwise.
\end{array}
\right.
\end{eqnarray*}

Step 4: Construct $\bigtriangleup f(U,\Delta,D)=(\{\mathscr{C}_{m+1}\})\bigwedge (\bigwedge_{x\in POS_{\cup\Delta^{+}}(D)\wedge x\notin \cup\mathscr{A}_{\mathscr{C}_{m+1}}}\bigvee r(x))$;

Step 5: Compute $\bigtriangleup\mathscr{R}(U,\Delta,D)=\{\Delta'_{j}\mid \overline{\exists}\Delta_{i}\in \mathscr{R}(U,\Delta,D), \text{ s.t. } \Delta_{i}\subset\Delta'_{j},1\leq j\leq k'\}$;

Step 6: Output $\mathscr{R}(U,\Delta^{+},D)=\mathscr{R}(U,\Delta,D)\cup (\triangle\mathscr{R}(U,\Delta,D))$.
\end{algorithm}

The time complexity of Step 3 is $[|U|\ast|\mathscr{C}_{m+1}|,|U|\ast|\mathscr{C}_{m+1}|\ast |U/D|]$; the time complexity of Steps 4 and 5 is $[|U|-|\cup\mathscr{A}_{\mathscr{C}_{m+1}}|,|U|\ast(|\Delta|+1)]$. Therefore, the
time complexity of the incremental algorithm is lower than that of
the non-incremental algorithm.

\begin{example}(Continuation from Example 4.2)
By Definition 2.6, we first have $
r(x_{1})=\{\mathscr{C}_{2},\mathscr{C}_{3}\},
r(x_{2})=\emptyset,
r(x_{3})=\emptyset,
r(x_{4})=\{\mathscr{C}_{1},\mathscr{C}_{2}\},
r(x_{5})=\{\mathscr{C}_{1},\mathscr{C}_{2}\},
r(x_{6})=\{\mathscr{C}_{1},\mathscr{C}_{2}\},
r(x_{7})=\{\mathscr{C}_{1}\},$ and $
r(x_{8})=\{\mathscr{C}_{1}\}.$
Thus we have
$R(U,\Delta,D)=\{\{\mathscr{C}_{2},\mathscr{C}_{3}\},
\{\mathscr{C}_{1},\mathscr{C}_{2}\},\{\mathscr{C}_{1}\}\}.
$
So we have
\begin{eqnarray*}
f(U,\Delta,D)&=&\bigwedge\{\bigvee r(x)\mid r(x)\in R(U,\Delta,D)\}\\
&=&(\mathscr{C}_{2}\vee\mathscr{C}_{3})\wedge(\mathscr{C}_{1}\vee\mathscr{C}_{2})\wedge
\mathscr{C}_{1}\\
&=&(\mathscr{C}_{2}\vee\mathscr{C}_{3})\wedge \mathscr{C}_{1}\\
&=&(\mathscr{C}_{1}\wedge\mathscr{C}_{2})\vee (\mathscr{C}_{1}\wedge\mathscr{C}_{3}).
\end{eqnarray*}
Thus, we have $\mathscr{R}(\Delta,U,D)=\{\{\mathscr{C}_{1},\mathscr{C}_{2}\}, \{\mathscr{C}_{1},\mathscr{C}_{3}\}\}.$

Secondly, by Definition 2.6, we have that
$r^{+}(x_{1})=\{\mathscr{C}_{2},\mathscr{C}_{3}\},
r^{+}(x_{2})=\emptyset,
r^{+}(x_{3})=\emptyset,
r^{+}(x_{4})=\{\mathscr{C}_{1},\mathscr{C}_{2},\mathscr{C}_{5}\},$ $
r^{+}(x_{5})=\{\mathscr{C}_{1},\mathscr{C}_{2},\mathscr{C}_{5}\},
r^{+}(x_{6})=\{\mathscr{C}_{1},\mathscr{C}_{2}\},
r^{+}(x_{7})=\{\mathscr{C}_{1}\},$ and $
r^{+}(x_{8})=\{\mathscr{C}_{1}\}.
$
So we have
\begin{eqnarray*}
f(U,\Delta^{+},D)&=&\bigwedge\{\bigvee r^{+}(x)\mid r^{+}(x)\in R(U,\Delta^{+},D)\}\\
&=&(\mathscr{C}_{1}\vee\mathscr{C}_{2})\wedge(\mathscr{C}_{2}\vee\mathscr{C}_{3})\wedge(\mathscr{C}_{1}\vee\mathscr{C}_{2}\vee\mathscr{C}_{5})\wedge
\mathscr{C}_{1}\\
&=&(\mathscr{C}_{2}\vee\mathscr{C}_{3})\wedge \mathscr{C}_{1}\\
&=&(\mathscr{C}_{1}\wedge\mathscr{C}_{2})\vee (\mathscr{C}_{1}\wedge\mathscr{C}_{3}).
\end{eqnarray*}
Thus, we have $\mathscr{R}(\Delta^{+},U,D)=\{\{\mathscr{C}_{1},\mathscr{C}_{2}\}, \{\mathscr{C}_{1},\mathscr{C}_{3}\}\}.$

Thirdly, by Theorem 4.4, we have $\mathscr{R}(\Delta^{+},U,D)=\{\{\mathscr{C}_{1},\mathscr{C}_{2}\}, \{\mathscr{C}_{1},\mathscr{C}_{3}\}\}$ since $POS_{\cup\Delta^{+}}(D)= POS_{\cup\Delta}(D)$.
\end{example}

Example 4.8 illustrates how to compute attribute reducts of $(\Delta^{+},D,U)$ by Algorithm 2.8; Example 4.8 also illustrates how to compute attribute reducts of $(\Delta^{+},D,U)$ by Algorithm 4.7. We see that the incremental algorithm is more effective than the non-incremental algorithm for attribute reduction of dynamic covering decision information systems.

In practical situations, there are a lot of dynamic covering information systems caused by deleting attributes, and we also study attribute reduction of dynamic covering information systems when deleting attributes as follows.

\begin{definition}
Let $(U,\Delta, D)$ and $(U,\Delta^{-}, D)$ be covering information systems, where $U=\{x_{1},x_{2},...,x_{n}\}$, $\Delta=\{\mathscr{C}_{1},
\mathscr{C}_{2},...,\mathscr{C}_{m-1},\mathscr{C}_{m}\}$, and $\Delta^{-}=\{\mathscr{C}_{1},
\mathscr{C}_{2},...,\mathscr{C}_{m-1}\}$. Then $(U,\Delta^{-}, D)$ is called a dynamic information system of $(U,\Delta, D)$.
\end{definition}

\noindent\textbf{Remark:} We take $(U,\Delta, D)$ as an inconsistent covering information system in Definition 4.9. We also notice that the dynamic covering information system $(U,\Delta^{-}, D)$ is inconsistent when deleting $\mathscr{C}_{m}$ from $(U,\Delta,D)$.

\begin{example}
Let $(U,\Delta, D)$ and $(U,\Delta^{-}, D)$ be covering information systems, where $U=\{x_{1},x_{2},...,x_{8}\}$, $\Delta=\{\mathscr{C}_{1},
\mathscr{C}_{2},\mathscr{C}_{3},\mathscr{C}_{4}\}$,  $\Delta^{-}=\{\mathscr{C}_{1},
\mathscr{C}_{2},\mathscr{C}_{3}\}$, and $U/D=\{\{x_{1},x_{2},x_{3}\},\{x_{4},x_{5},x_{6}\},\{x_{7},x_{8}\}\}$, where
\begin{eqnarray*}
\mathscr{C}_{1}&=&\{\{x_{1},x_{2},x_{3},x_{4}\},\{x_{3},x_{6},x_{7}\},
\{x_{4},x_{5}\},\{x_{6}\},\{x_{7},x_{8}\}\};\\
\mathscr{C}_{2}&=&\{\{x_{1}\},\{x_{2},x_{3},x_{4}\},\{x_{4},x_{5}\},
\{x_{4},x_{5},x_{6}\},\{x_{6},x_{7},x_{8}\}\};\\
\mathscr{C}_{3}&=&\{\{x_{1}\},\{x_{1},x_{3},x_{4}\},\{x_{2},x_{3},x_{4},x_{8}\},
\{x_{3},x_{4},x_{5},x_{6},x_{7}\}\};\\
\mathscr{C}_{4}&=&\{\{x_{1},x_{4},x_{5}\},\{x_{2},x_{3},x_{4},x_{5}\},
\{x_{4},x_{5},x_{6},x_{7},x_{8}\}\}.
\end{eqnarray*}
By Definition 4.9, we see that $(U,\Delta^{-},D)$ is a dynamic information system of $(U,\Delta,D)$. Specially, $(U,\Delta,D)$ and $(U,\Delta^{-},D)$ are
inconsistent covering information systems.
\end{example}

Suppose $(U,\Delta^{-}, D)$ and $(U,\Delta, D)$ are covering information systems, where $U=\{x_{1},x_{2},...,x_{n}\}$, $\Delta=\{\mathscr{C}_{1},\mathscr{C}_{2},...,\mathscr{C}_{m}\}$, and $\Delta^{-}=\{\mathscr{C}_{1},\mathscr{C}_{2},...,\mathscr{C}_{m-1}\}$,
$\mathscr{A}_{\Delta}=\{C_{k}\in \cup \Delta\mid \exists D_{j}\in U/D, \text{ s.t. } C_{k}\subseteq D_{j}\}$, $r(x)=\{\mathscr{C}\in \Delta\mid \exists C_{k}\in \mathscr{A}_{\Delta}, \text{ s.t. } x\in C_{k}\in \mathscr{C}\},$ and  $r^{-}(x)=\{\mathscr{C}\in \Delta^{-}\mid \exists C_{k}\in \mathscr{A}_{\Delta^{-}}, \text{ s.t. } x\in C_{k}\in \mathscr{C}\}.$

\begin{theorem}
Let $(U,\Delta, D)$ and $(U,\Delta^{-}, D)$ be covering information systems, where $U=\{x_{1},x_{2},...,x_{n}\}$, $\Delta=\{\mathscr{C}_{1},
\mathscr{C}_{2},...,\mathscr{C}_{m}\}$, and $\Delta^{-}=\{\mathscr{C}_{1},
\mathscr{C}_{2},...,\mathscr{C}_{m-1}\}$. Then we have
\makeatother $$r^{-}(x)=\left\{
\begin{array}{ccc}
r(x)\backslash\{\mathscr{C}_{m}\},&{\rm if}& x\in \cup\mathscr{A}_{\mathscr{C}_{m}};\\
r(x),&{\rm }& otherwise.
\end{array}
\right. $$
\end{theorem}

\noindent\textbf{Proof:} The proof is similar to Theorem 3.4.$\Box$

Theorem 4.11 illustrates the relationship between $r(x)$ of $(U,\Delta, D)$ and $r^{-}(x)$ of $(U,\Delta^{-}, D)$, which reduces the time complexities of computing related family $R(U,\Delta^{-},D)$.

\begin{theorem}
Let $(U,\Delta, D)$ and $(U,\Delta^{-}, D)$ be covering information systems, where $U=\{x_{1},x_{2},...,$ $x_{n}\}$, $\Delta=\{\mathscr{C}_{1},\mathscr{C}_{2},...,\mathscr{C}_{m}\}$, and $\Delta^{-}=\{\mathscr{C}_{1},\mathscr{C}_{2},...,\mathscr{C}_{m-1}\}$. If $POS_{\cup\Delta^{-}}(D)=POS_{\cup\Delta}(D)$, then
we have $\mathscr{R}(U,\Delta^{-},D)=\{\Delta_{i}\mid \mathscr{C}_{m}\notin\Delta_{i}\in \mathscr{R}(U,\Delta,D)\}.$
\end{theorem}

\noindent\textbf{Proof:} The proof is straightforward by Definition 2.6.$\Box$

Theorem 4.12 illustrates the relationship between $\mathscr{R}(U,\Delta^{-},D)$ and $\mathscr{R}(U,\Delta,D)$, and we get $\mathscr{R}(U,\Delta^{-},D)$ $=\{\Delta_{i}\mid \mathscr{C}_{m}\notin\Delta_{i}\in \mathscr{R}(U,\Delta,D)\}$ when $POS_{\cup\Delta^{-}}(D)=POS_{\cup\Delta}(D)$, which reduces the time complexities of computing attribute reducts of dynamic covering information systems.

\begin{theorem}
Let $(U,\Delta, D)$ and $(U,\Delta^{-}, D)$ be covering information systems, where $U=\{x_{1},x_{2},...,$ $x_{n}\}$, $\Delta=\{\mathscr{C}_{1},\mathscr{C}_{2},...,\mathscr{C}_{m}\}$, and $\Delta^{-}=\{\mathscr{C}_{1},\mathscr{C}_{2},...,\mathscr{C}_{m-1}\}$. If $POS_{\cup\Delta^{-}}(D)\neq POS_{\cup\Delta}(D)$, then
we have $\mathscr{R}(U,\Delta^{-},D)=\{\Delta_{i}\backslash \{\mathscr{C}_{m}\}\mid \Delta_{i}\in \mathscr{R}(U,\Delta,D)\}.$
\end{theorem}

\noindent\textbf{Proof:} The proof is straightforward by Definition 2.6.$\Box$

Theorem 4.13 illustrates the relationship between $\mathscr{R}(U,\Delta^{-},D)$ and $\mathscr{R}(U,\Delta,D)$, and we get $\mathscr{R}(U,\Delta^{-},D)$ $=\{\Delta_{i}\backslash \{\mathscr{C}_{m}\}\mid \Delta_{i}\in \mathscr{R}(U,\Delta,D)\}$ when $POS_{\cup\Delta^{-}}(D)\neq POS_{\cup\Delta}(D)$, which reduces the time complexities of computing attribute reducts of dynamic covering information systems.

We provide incremental algorithm of computing $\mathscr{R}(U,\Delta^{-},D)$ for consistent covering information system $(U,\Delta^{-},D)$ as follows.

\begin{algorithm}(Incremental Algorithm of Computing $\mathscr{R}(U,\Delta^{-},D)$ for Covering Information System $(U,\Delta^{-},D)$)(IAIDIS)

Step 1: Input $(U,\Delta^{-}, D)$;

Step 2: Construct $POS_{\cup\Delta^{-}}(D)$;

Step 3: Compute $\mathscr{R}(U,\Delta^{-},D)=\{\Delta_{i}\mid \mathscr{C}_{m}\notin\Delta_{i}\in \mathscr{R}(U,\Delta,D)\}$ when $POS_{\cup\Delta^{-}}(D)=POS_{\cup\Delta}(D)$;

Step 4: Construct $\mathscr{R}(U,\Delta^{-},D)=\{\Delta_{i}\backslash \{\mathscr{C}_{m}\}\mid \Delta_{i}\in \mathscr{R}(U,\Delta,D)\}$ when $POS_{\cup\Delta^{-}}(D)\neq POS_{\cup\Delta}(D)$;

Step 5: Output $\mathscr{R}(\Delta^{-},U,D)$.
\end{algorithm}

The time complexity of Step 2 is $[|U|\ast(\sum_{\mathscr{C}\in\Delta^{-}}|\mathscr{C}|),|U|\ast(\sum_{\mathscr{C}\in\Delta^{-}}|\mathscr{C}|)\ast |U/D|]$; the time complexity of Step 3 is $[|U|^{2},|U|^{2}\ast(\sum_{\mathscr{C}\in\Delta}|\mathscr{C}|)\ast |U/D|]$; the time complexity of Steps 3 and 4 is $|\mathscr{R}(U,\Delta,D)|$. Therefore, the
time complexity of the non-incremental algorithm is very high. Therefore, the
time complexity of the incremental algorithm is lower than that of
the non-incremental algorithm.

\begin{example}(Continuation from Example 4.10)
By Definition 2.6, we first have $
r^{-}(x_{1})=\{\mathscr{C}_{2},\mathscr{C}_{3}\},
r^{-}(x_{2})$ $=\emptyset,
r^{-}(x_{3})=\emptyset,
r^{-}(x_{4})=\{\mathscr{C}_{1},\mathscr{C}_{2}\},
r^{-}(x_{5})=\{\mathscr{C}_{1},\mathscr{C}_{2}\},
r^{-}(x_{6})=\{\mathscr{C}_{1},\mathscr{C}_{2}\},
r^{-}(x_{7})=\{\mathscr{C}_{1}\},$ and $
r^{-}(x_{8})=\{\mathscr{C}_{1}\}.$
So we have
\begin{eqnarray*}
f(U,\Delta^{-},D)&=&\bigwedge\{\bigvee r^{-}(x)\mid r^{-}(x)\in R(U,\Delta,D)\}\\
&=&(\mathscr{C}_{2}\vee\mathscr{C}_{3})\wedge(\mathscr{C}_{1}\vee\mathscr{C}_{2})\wedge
\mathscr{C}_{1}\\
&=&(\mathscr{C}_{2}\vee\mathscr{C}_{3})\wedge \mathscr{C}_{1}\\
&=&(\mathscr{C}_{1}\wedge\mathscr{C}_{2})\vee (\mathscr{C}_{1}\wedge\mathscr{C}_{3}).
\end{eqnarray*}
Thus, we have $\mathscr{R}(U,\Delta^{-},D)=\{\{\mathscr{C}_{1},\mathscr{C}_{2}\}, \{\mathscr{C}_{1},\mathscr{C}_{3}\}\}.$

Secondly, by Theorem 4.12, we have $\mathscr{R}(U,\Delta^{-},D)=\{\{\mathscr{C}_{1},\mathscr{C}_{2}\}, \{\mathscr{C}_{1},\mathscr{C}_{3}\}\}$ since $POS_{\cup\Delta^{-}}(D)=POS_{\cup\Delta}(D)$.
\end{example}

Example 4.15 illustrates how to compute attribute reducts of $(\Delta^{-},D,U)$ by Algorithm 2.8; Example 4.15 also illustrates how to compute attribute reducts of $(\Delta^{-},D,U)$ by Algorithm 4.14. We see that the incremental algorithm is more effective than the non-incremental algorithm for attribute reduction of dynamic covering decision information systems.

%
%

\section{Conclusions}

In this paper, we have constructed attribute reducts of consistent covering information systems with variations of attribute sets. We have employed examples to illustrate how to compute attribute reducts of consistent covering information systems when varying attribute sets.
Furthermore, we have investigated updated mechanisms for constructing attribute reducts of inconsistent covering information systems with variations of attribute sets. We have employed examples to illustrate how to compute attribute reducts of inconsistent covering information systems when varying attribute sets.
Finally, we have employed the experimental results to illustrate that the related family-based incremental approaches are effective for attribute reduction of dynamic covering information systems when attribute sets are varying with time.

\section*{ Acknowledgments}

We would like to thank the anonymous reviewers very much for their
professional comments and valuable suggestions. This work is
supported by the National Natural Science Foundation of China (NO.61673301, 61603063, 11526039, 61573255), Doctoral Fund of Ministry of Education of China(No. 20130072130004), China Postdoctoral Science Foundation(NO.2013M542558, 2015M580353), the Scientific
Research Fund of Hunan Provincial Education Department(No.15B004).

\end{document}